\documentclass[11pt,a4paper]{article}
\usepackage[hyperref]{acl2020}
\usepackage{times}
\usepackage{latexsym}

\usepackage{microtype}
\usepackage{rotating}

\usepackage{tabularx}
\usepackage{array, caption, multirow, tabularx,  ragged2e,  booktabs}
\usepackage{multirow}
\usepackage{multicol}
\usepackage{tabularx}
\usepackage[linesnumbered,ruled,vlined]{algorithm2e}

\SetCommentSty{mycommfont}
\usepackage{tikzsymbols}
\usepackage{pifont}
\newcommand{\xmark}{\ding{55}}%
\usepackage[scriptsize]{subfigure}
\usepackage{amsmath,amsfonts,amssymb,amscd,amsthm,xspace}
\SetKwInput{KwInput}{Input}                
\SetKwInput{KwOutput}{Output}              

\usepackage{algpseudocode}
\aclfinalcopy 

\title{Automated Evidence Collection for Fake News Detection}

\author{Mrinal Rawat \\
  UpGrad Education Pvt. Ltd., Mumbai, India.\\
  Liverpool John Moores University, \\
  Liverpool, United Kingdom. \\
  \texttt{rawatmrinal06@gmail.com} \\\And
  Diptesh Kanojia \\
  Centre for Translation Studies, \\
  University of Surrey, \\
  Surrey, United Kingdom. \\
  \texttt{d.kanojia@surrey.ac.uk} \\}

\date{29-11-2021}

\begin{document}
\maketitle
\begin{abstract}
Fake news, misinformation, and unverifiable facts on social media platforms propagate disharmony and affect society, especially when dealing with an epidemic like COVID-19. The task of Fake News Detection aims to tackle the effects of such misinformation by classifying news items as fake or real. In this paper, we propose a novel approach that improves over the current automatic fake news detection approaches by automatically gathering evidence for each claim. Our approach extracts supporting evidence from the web articles and then selects appropriate text to be treated as evidence sets. We use a pre-trained summarizer on these evidence sets and then use the extracted summary as supporting evidence to aid the classification task. Our experiments, using both machine learning and deep learning-based methods, help perform an extensive evaluation of our approach. The results show that our approach outperforms the state-of-the-art methods in fake news detection to achieve an F1-score of 99.25 over the dataset provided for the CONSTRAINT-2021 Shared Task. We also release the augmented dataset, our code and models~\footnote{\url{https://github.com/rawat-mrinal06/fake_news}} for any further research.
\end{abstract}

\section{Introduction}
The ability to consume readily available information from the internet is alarming for both individuals and organizations. The quality of content on social media platforms has been significantly affected due to the spread of fake news, misinformation and unverifiable facts. The current tally of internet users stands at 4.66 billion\footnote{\href{https://www.internetlivestats.com/}{Internet Live Stats (as on 15-07-2021)}}~\citep{kemp2015global}; and many of these users generate, post and consume content without any regulation, in a large number of countries\footnote{\href{https://en.wikipedia.org/wiki/Internet_censorship_and_surveillance_by_country}{Internet Censorship in Countries}}. Due to the unrestricted nature of online platforms, there is a significant increase in the amount of misinformation on social media~\citep{allen2020evaluating}, especially in developing nations~\citep{badrinathan2020educative,wasserman2019exploratory}. Studies show that events such as the presidential election of the United States in 2016 were affected due to \textit{moderated fake news} campaigns~\citep{tavernise2016fake}. Shu et al.(2017)~\cite{shu2017fake} propose that \textit{fake news is intentionally written, verifiably false, and is created in a way that makes it look authentic}. Manual efforts by other online platforms such as Poynter\footnote{\href{https://www.poynter.org/}{Poynter: Online}}, FactCheck\footnote{\href{https://www.factchecker.in/}{FactCheck: Online}}, AltNews\footnote{\href{https://www.altnews.in/}{AltNews: Online}} \textit{etc.} to detect fake news, requires a lot of human effort and can prove to be cumbersome. Such manual efforts can be time-consuming, challenging, and at times, can also be ineffective as fake news can spread faster than verified claims over social media platforms.

Automatic Fake News Detection is a task that aims to mitigate the problem of misinformation with the help of evidence supported by various sources. Most of the approaches in this recently devised task aim to use the classical machine learning-based methods or the recent deep learning-based methods to help classify news items as fake or as real. Initially proposed methods for the task applied machine learning-based techniques but cited insufficient data as a major concern~\citep{vlachos-riedel-2014-fact}. Recent deep learning and ensemble approaches \citep{malon-2018-team,DBLP:journals/corr/abs-1811-04670} were proposed on the FEVER \citep{thorne-etal-2018-fever} and LIAR \citep{wang-2017-liar} datasets, and have been shown to perform very well. Studies have proposed a combination of evidence detection with textual entailment concerning the claim~\citep{vijjali-etal-2020-two}. FEVER Shared Tasks~\cite{thorne2018fact,thorne-etal-2019-fever2} have helped the automatic fact verification task gather attention towards the problem and helped generate approaches to mitigate the issues with previously proposed solutions. This study shows that our novel approach improvises over state-of-the-art approaches and helps detect fake news related to COVID-19. Our approach performs web-search for evidence collection and uses BERT-score similarity to match the unverified claim with the top-k searches. Further, we propose the use of summarization to mitigate problems with the evidence collection. We summarize the top-n selected lines from these articles and use them as evidence to support or reject the news item claim. Our experiments perform an extensive evaluation of the approach over the datasets released as a part of the CONSTRAINT-2021 Shared Task~\citep{patwa2020fighting} shared task.\\
\\
Our summarized contributions with this paper are:

\begin{itemize}
    \item We propose a novel approach to help automate the evidence collection for any fake news detection dataset.
    \item Additionally, we incorporate a summarization component that helps outperform the state-of-the-art approaches for automatic fact verification on the CONSTRAINT-2021 dataset.
\end{itemize}

\section{Related Work}
Automatic detection and classification of fake news, especially in epidemic situations like COVID-19, is a significant issue for society. Most of the recent works have identified that fake news is written intentionally and factually false~\citep{shu2017fake}. Several datasets have been released for the AI community in the field of fake news detection, such as LIAR~\citep{wang-2017-liar}, Fake News Challenge-1, and FEVER~\citep{thorne-etal-2018-fever}. Some recent techniques extract the evidence from Wikipedia to classify a claim as  \texttt{SUPPORTED, REFUTED or NOTENOUGHINFO}~\citep{thorne-etal-2018-fever}. They formulate the problem as a three-step process (i) first the top-k documents are identified based on the TF-IDF based approaches (ii) then top-k sentences are identified from the documents, and (iii) finally the textual entailment based approaches~\citep{DBLP:journals/corr/ParikhT0U16} are used to classify the claim. Team Papelo~\citep{malon-2018-team} used the Transformer-based approach for the textual entailment and selected the evidence-based on tf-idf and entities present in the title. \citet{hanselowski-etal-2018-ukp} selects the documents and sentence using the entity mentions and recognizes textual entailment using Enhanced Sequential Inference Model (ESIM) \citep{chen-etal-2017-enhanced}. Despite the several attempts, fake news detection is a challenging problem and countering fake news is a typical issue that requires continuous studies. Recently, some researchers released the datasets related to COVID-19 fake news detection. Shahi and Nandini (2020) \cite{DBLP:journals/corr/abs-2006-11343} proposed the first multi-lingual cross-domain dataset for COVID-19 that consists of 5182 fact-checked news articles from Jan-2020 to May-2020. They collected data from 92 different websites and manually classified them into 23 classes. Kar et al. \cite{DBLP:journals/corr/abs-2010-06906} also released a multi-indic-lingual dataset besides English to detect the fake news in social media tweets. They obtained 480 tweets in Bengali and 460 tweets in Hindi. In addition to tweets, they also included several features related to tweets such as retweet count, favourite count, total URL in description, URL, friend list, followers, \textit{etc.} Recently, a very relevant dataset was released by Patwa et al. \cite{patwa2020fighting} which consists of 10,700 tweets or claim collected from various sources such as Twitter, PolitiFact, Snopes, Boomlive. They experimented with various machine learning techniques like Decision Trees, Logistic Regression, SVM, Gradient Boosting DT and achieved the F1-score of 93.32. Most of the previous work on COVID-19 dataset proposed an ensemble approach of various models such as BERT, RoBERTa, XLNet, \textit{etc.}~\cite{DBLP:journals/corr/abs-2101-12027, DBLP:journals/corr/abs-2101-11954, 2021arXiv210105701S}. Chen et al.~\citep{10.1007/978-3-030-73696-5_9} trained the model with additional words such as covid-19, coronavirus, pandemic, indiafightscorona since the BERT tokenizer will split these words into separate tokens. Some works leveraged the fine-tuned models like COVID-Twitter-BERT (CT-BERT) \citep{DBLP:journals/corr/abs-2005-07503} and demonstrated a boost in performance \citep{10.1007/978-3-030-73696-5_11,DBLP:journals/corr/abs-2012-11967,DBLP:journals/corr/abs-2101-04012}. The fake news detection methods described above mainly uses the claim for the classification. Our method focuses on extracting and summarizing the evidence from the external source and uses it to classify the claim on the COVID-19 fake news detection dataset~\citep{10.1007/978-3-030-73696-5_11,patwa2020fighting}.

\begin{table}[h!]
\centering
\begin{tabular}{|l|c|c|r|} \hline
\textbf{Split} & \textbf{Real} & \textbf{Fake} & \textbf{Total} \\ \hline
Training & 3360 & 3060 & 6420  \\\hline
Validation & 1120 & 1020 & 2140 \\\hline
Test & 1120 & 1020 & 2140 \\\hline
\textbf{Total} & 5600 & 5100 & 10700 \\\hline
\end{tabular}
\caption{Dataset Statistics}
\label{table:stats}
\vspace{-0.7cm}
\end{table}

\section{Dataset}
\label{section:dataset}

For our work, we use the pre-released COVID-19 fake news dataset as a part of the CONSTRAINT-2021 shared task~\citep{patwa2020fighting}. This gold-standard manually annotated dataset comprises social media posts and articles which are related to COVID-19. Each post or tweet contains content in the English language and is classified in either of the two categories- \textbf{(1) Real:} where tweets or articles which are factually correct and verified from authentic sources, for example, ``Wearing mask can protect you from the virus. (Twitter)''; or \textbf{(2) Fake:} where tweets or posts related to COVID-19 which are factually incorrect and verified as false, for example, ``If you take Crocin thrice a day you are safe. (Facebook)''.

The authors collect fake news from two different sources- social media platforms and public fact-checking platforms. The social media posts include text from Facebook posts, Instagram posts, and Twitter posts, whereas the fact-checking websites such as PolitiFact, Snopes, and Boomlive are used to collect fact-checked news items. To further collect real news, they sample tweets from official government channels, news channels, and medical institutes. Overall, a total of 14 such sources were used to prepare this dataset. 

\begin{figure}[!t]
\centering
\subfigure[Worcloud of Real Posts]{
\includegraphics[width=.40\columnwidth]{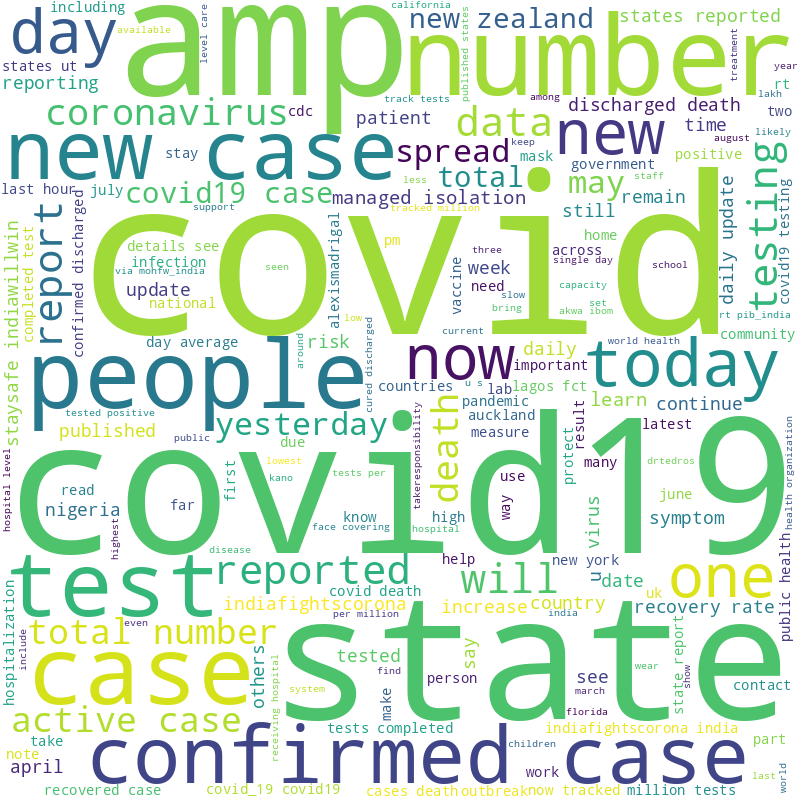}
}
\subfigure[Worcloud of Fake Posts]{
\includegraphics[width=.40\columnwidth]{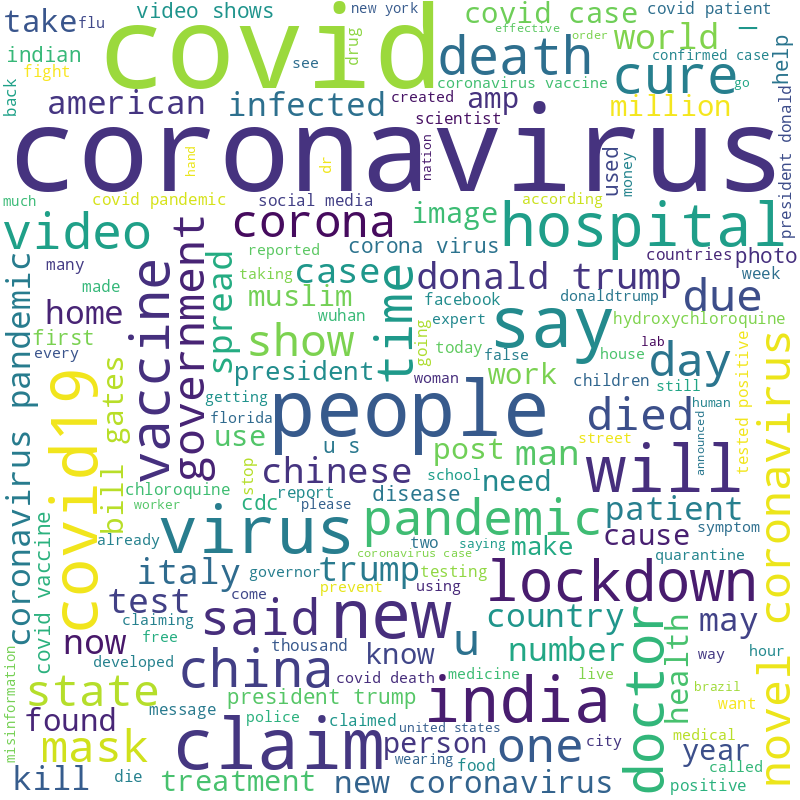}
}
\caption{Wordcloud of real/fake posts in our dataset.}
\label{fig:wordcloud}
\vspace{-0.3cm}
\end{figure}

The dataset comprises 10700 manually annotated samples and is split into (60\%) train, (20\%) validation and (20\%) test sets. We provide the exact numbers for each split/class in Table~\ref{table:stats} for clarity. The dataset is class-balanced as it contains 52.3\% samples of real posts and 47.7\% samples of fake posts. As an analysis on it, we obtained a word-cloud illustration for both real and fake samples and observed a high lexical overlap between both the classes, where words like `\textit{coronavirus}', `\textit{covid19}', `\textit{people}', `\textit{cases}', `\textit{number}', `\textit{test}', \textit{etc.} are repeatedly used in both the sets. We do not show the word cloud due to space constraints. We create and present a wordcloud for the dataset in Figure~\ref{fig:wordcloud}.

\section{Our Approach}
In this section, we provide details of our novel approach to augment the dataset with evidence from web search and the use of this evidence to complement the task of fact verification. The algorithm for our approach can be seen in Algorithm~\ref{algo}.
\begin{algorithm}[]
\DontPrintSemicolon
  \KwInput{Claim $c$, Blocked URLs $u$}
  \KwOutput{Evidence $e$}

\SetKwFunction{SearchF}{ArticleRet}
  \SetKwProg{Fn}{Function}{:}{}
  \Fn{\SearchF{$c$, $k=3$}}{
        
        $results \gets GoogleSearch($c$);$
        $filtered\_results \gets \emptyset$;
        \ForEach{$r_i \in results$}{%
            \If{$r_i \notin u$}{
                     
                $s_i \gets$ Similarity($c$, $r_i$); \tcp*{Document Similarity using spacy library}
                \If{$s_i > 0.7$}{
                    $filtered\_results \gets (r_i, s_i)$
                }
            }
        }
        $filtered\_results \gets Sort(filtered\_results)[:$k$];$
        $\KwRet\ filtered\_results;$
  }
  $articles$ $\gets \SearchF{c}$;
  
  $e \gets \emptyset$;   \tcp*{Evidences}
  \ForEach{$a_i \in articles$}{%
        $d \gets WebsiteData(a_i.url)$; 
        $sents \gets d['<h>'] + d['<p>']$;  \tcp*{Extract <p> and <h> tags from html}
        \ForEach{$s_i \in sents$}{%
            $sim \gets Similarity(c, s_i);$ 
            \If{$sim > 0.5$}{
                    $e$ $\gets$ ($s_i$, $sim$);
            }
        }
        $e$ $\gets Sort($e$)[:$3$];$
  }
 $\KwRet$ $e;$

\caption{Algorithm to collect the evidence from the input claim}
\label{algo}
\end{algorithm}
As discussed above, we collect this evidence and prune to top-k related news items based on semantic similarity via BERTScore~\citep{devlin-etal-2019-bert}. We also select top-n lines from each article for further building an evidence repository, as detailed below in further subsections.

\begin{figure*}[t!]
\centering
\includegraphics[width=1\textwidth]{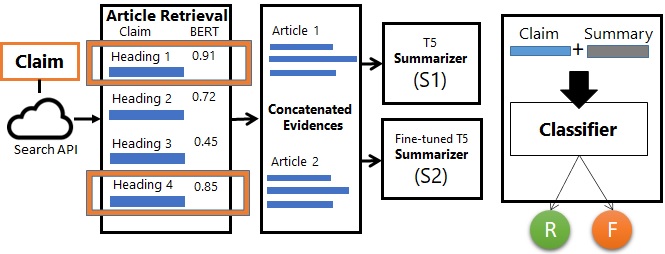}
\caption{Full architecture of our proposed approach.}
\label{fig:arch}
\end{figure*}

\subsection{Evidence Collection}
In the original dataset of COVID-19, evidence is not released along with the claim. We hypothesize that evidence is equally relevant to classify the claim as proposed by~\citet{thorne-etal-2018-fever}. As per our approach, given a claim or post text, we first select K relevant articles using a BERT-based sentence similarity score as detailed here; and can be seen as an architecture component in Figure~\ref{fig:arch}.

\subsubsection{Article Retrieval}
For each claim $c$, we search the claim as a query using a publicly available search API. The response returned by this API consists of \texttt{(heading, text)} pairs. We use the spacy~\cite{spacy} library to get the similarity score of response text with respect to the input claim. Based on this similarity score, we select top K results that have the similarity score greater than $0.7$\footnote{Selected with empirical evaluation and manual analysis after trying 0.5, 0.6, 0.7, 0.8 using the 'en\_nli\_roberta\_base' model for document similarity}. While selecting documents, we prune for webpages in other languages and pages which are direct links to PDF or other such non-text files. As an immediate next step, we scrape the selected web pages to obtain the matching N sentences concerning the claim as detailed here.

\subsubsection{Sentence(s) Retrieval}
In the previous step, we extract the relevant article URLs $U = (u_1, u_2, u_3)$. We employ a similar method to find the sentences within each article. For every url $u$, we first scrape the webpage and extract the text from $<h>$ and $<p>$ tags. Further, we use the same similarity score to select the top N sentences with respect to the claim. We obtain a similarity threshold of 0.5 after performing a similar empirical evaluation as mentioned in the footnote. Eventually, we concatenate the selected sentences from these articles, which act as our evidence for the claim. We would like to note that increasing the threshold significantly higher returned an empty set in some case and hence we choose a relatively lower threshold (0.5).

An example of evidence collected via our approach is shown in Table~\ref{tab:summary} where the column titled ``Evidence'' shows the output after these steps.

\begin{table*}[t]
\centering
\resizebox{\textwidth}{!}{%
\begin{tabular}{@{}llll@{}}
\toprule
\multicolumn{1}{c}{\textbf{Claim}} & \multicolumn{1}{c}{\textbf{Evidence}} & \multicolumn{1}{c}{\textbf{Summarization-1 (S1)}} & \multicolumn{1}{c}{\textbf{Summarization-2 (S2)}} \\ \midrule
\begin{tabular}[c]{@{}l@{}}There is no evidence that children have \\ died because of a COVID-19 vaccine. \\ No vaccine currently in development has \\ been approved for widespread public use.\\ https://t.co/9ecvMR8SAf\end{tabular} & \begin{tabular}[c]{@{}l@{}}Currently there is no coronavirus vaccine that has been\\ approved for the American public. And there is no \\ evidence that children have died because they received \\ one of the COVID-19 vaccines being developed.\\ PolitiFact found no evidence that anyone has died from\\ complications related to a trial COVID-19 vaccination.\\ There is no evidence that children have died because \\ of a COVID-19 vaccine.\end{tabular} & \begin{tabular}[c]{@{}l@{}}There is no evidence that children have died\\ because they received a COVID-19 vaccine.\\ No evidence that anyone has died from \\ complications related to a trial COVID-19.\end{tabular} & \begin{tabular}[c]{@{}l@{}}There is no evidence that children have \\ died because they received one of the \\ COVID-19 vaccines being developed.\\ PolitiFact found no evidence that anyone\\ has died from complications related to a \\ trial COVID-19 vaccination.\end{tabular} \\ \bottomrule
\end{tabular}%
}
\caption{Illustrative example of our approach pipeline shown as Claim $\xrightarrow{}$ Evidence $\xrightarrow{}$ Summarization-1 $\xrightarrow{}$ Summarization-2, where Summarization-2 is obtained after fine-tuning T5 language model, and used as \textbf{evidence input} for classification}
\label{tab:summary}
\end{table*}

\subsection{Dataset Preprocessing}
To map claims with evidence, we pre-process both the dataset and the evidence collected from external sources. Following are the details of the pre-processing steps: \textbf{(1) URL Mapping:} We observe that some posts contain URLs in a masked form, \textit{e.g.,} https://t.co/z5kk XpqkYb. Our approach extracts these URLs using a regular expression-based match and maps them to the original URL using the python `\textit{requests}' library. Any additional information from the URL is removed, and only appropriate URLs remain in the text. For example, https://t.co/z5kkXpqkYb  $\xrightarrow{}$  https://www.cdc.gov/; \textbf{(2) Special symbols:} We removed extra white-spaces, special symbols and brackets like \^ , (, ),  \{,\}; \textbf{(3) Hashtags, Emojis and Mentions:} Additionally, we remove hashtags and replace it with the token ``HASHTAG:''. 

For example, \#COVID-19 becomes HASHTAG:COVID-19. Similarly, we also replace mentions ``@'' with ``MENTION:'' token. At the end, we convert emojis to their text form using the `demoji' library\footnote{\href{https://github.com/bsolomon1124/demoji
}{GitHub: Demoji}}; \textbf{(4) Lowercasing:} Eventually, we lowercase the claim and the evidence text to obtain the input data used for the next Summarization step.

\subsection{Summarization of Evidence}
Our pre-processed evidence for claims in many cases were multiple paragraphs resulting in performance degradation. Therefore, we propose the addition of a summarization component to our pipeline which utilizes state-of-the-art Text-to-Text Transfer Transformer (T5) language model~\citep{DBLP:journals/corr/abs-1910-10683} for the inherent summarization task\footnote{This language model can perform the summarization task with the help of a prefix ``summarize'' to the input text provided.}. Due to the nature of the usual summarization task input, a large body of text (full documents), we believe that our comparatively short paragraphs would be better summarized. This language model is fine-tuned for the task of summarization helps us obtain a summarized text for each piece of evidence resulting in what we call Summarization-1 or S1. An output obtained is shown in Table~\ref{tab:summary}.

\subsubsection{Fine-tuning T5 on FEVER Dataset}
As an additional experimental step, we further fine-tune the Text-to-Text Transfer Transformer (T5) model using the original FEVER dataset~\citep{thorne-etal-2018-fever}. The original T5 summarization model is trained on the CNN/Daily Mail~\cite{DBLP:conf/nips/HermannKGEKSB15} data where the input is the news article text, and the objective is to highlight summarized text as the output. The T5 is an encoder-decoder model pre-trained on a multi-task mixture of unsupervised and supervised tasks and for which each task is converted into a text-to-text format. This allows for the use of the same model, loss function, hyperparameters, \textit{etc.} across our diverse set of tasks. T5 works well on a variety of tasks out-of-the-box by prepending a different prefix to the input corresponding to each task, e.g., for the task of translation$\xrightarrow{}$ \textit{translate English to German: $<$English Sentence$>$}, for the task of summarization$\xrightarrow{}$ \textit{summarize: $<$English Text$>$}.

For our experiments, the aim is to summarize the pre-processed evidence while including the claim. Thus, we hypothesize that fine-tuning on an auxiliary dataset will improve the quality of the generated summary. For fine-tuning, we use the same hyperparameters as described in their paper to generate another model. We perform another iteration of the summarization step using this fine-tuned model to generate a parallel set of evidence and label the output as Summarization-2 or S2 as shown in Table~\ref{tab:summary}. Further, we provide the details of \textbf{as the classification task, which uses either S1 or S2 as evidence input to classify} the claims as real or fake (Figure~\ref{fig:arch}). 

\begin{table*}[t!]
\centering
\resizebox{\textwidth}{!}{%
\begin{tabular}{|l|c|c|l|l|l|l|l|l|l|l|l|l|l|l|l|l|l|l|l|l|}
\hline
 & \multicolumn{2}{c|}{Previous Approaches} & \multicolumn{9}{c|}{\textbf{Our Approach w/ various Classification methods}} \\ \hline
\multirow{2}{*}{} & \multirow{2}{*}{\begin{tabular}[c]{@{}c@{}}~\citet{10.1007/978-3-030-73696-5_9}\end{tabular}} & \multirow{2}{*}{\begin{tabular}[c]{@{}c@{}}~\citet{10.1007/978-3-030-73696-5_11}\end{tabular}} & \multicolumn{3}{c|}{Logistic Regression} & \multicolumn{3}{c|}{SVM} & \multicolumn{3}{c|}{LSTM} \\ \cline{4-12} 
 &  &  & \multicolumn{1}{c|}{-} & \multicolumn{1}{c|}{S1} & \multicolumn{1}{c|}{S2} & \multicolumn{1}{c|}{-} & \multicolumn{1}{c|}{S1} & \multicolumn{1}{c|}{S2} & \multicolumn{1}{c|}{-} & \multicolumn{1}{c|}{S1} & \multicolumn{1}{c|}{S2} \\ \hline
\textbf{P} & 0.9902 & 0.986 & 0.9531 & 0.9565 & 0.9701 & 0.9641 & 0.9671 & 0.9764 & 0.9589 & 0.9598 & 0.9612  \\ \hline
\textbf{R} & 0.9901 & 0.985 & 0.9531 & 0.9564 & 0.9700 & 0.9639 & 0.9668 & 0.9761 & 0.9584 & 0.9596 & 0.9612  \\ \hline
\textbf{F} & 0.9901 & 0.985 & 0.9531 & 0.9565 & \underline{0.9700} & 0.9639 & 0.9668 & \underline{0.9761} & 0.9584 & 0.9596 & \underline{0.9612}  \\ \hline
\end{tabular}%
}
\caption{Results obtained after the fake news classification task where the values for previous approaches are from the latest shared task results and the results for each iteration of our approach are shown [P (Precision), R (Recall), and F (F-Score)]. (-) $\xrightarrow{}$ No Evidence, S1 $\xrightarrow{}$ Summarization-1 as Evidence, S2 $\xrightarrow{}$ Summarization-2 as Evidence.}
\label{tab:results}
\end{table*}

\begin{table*}[t!]
\centering
\resizebox{\textwidth}{!}{%
\begin{tabular}{|l|l|l|l|l|l|l|l|l|l|}
\hline
 & \multicolumn{9}{c|}{\textbf{Our Approach w/ various Deep Learning Classification methods}} \\ \hline
\multirow{2}{*}{} & \multicolumn{3}{c|}{BERT$_{base}$} & \multicolumn{3}{c|}{RoBERTa$_{base}$} & \multicolumn{3}{c|}{XLNet$_{base}$} \\ \hline 
 &  \multicolumn{1}{c|}{-} & \multicolumn{1}{c|}{S1} & \multicolumn{1}{c|}{S2} & \multicolumn{1}{c|}{-} & \multicolumn{1}{c|}{S1} & \multicolumn{1}{c|}{S2} & \multicolumn{1}{c|}{-} & \multicolumn{1}{c|}{S1} & \multicolumn{1}{c|}{S2} \\ \hline
\textbf{P} & 0.9612 & 0.9916 & 0.9917 & 0.9918 & 0.9929 & 0.9922 & 0.9920 & 0.9934 & 0.9947 \\ \hline
\textbf{R} & 0.9864 & 0.9888 & 0.9897 & 0.9897 & 0.9911 & 0.9916 & 0.9892 & 0.9911 & 0.9925 \\ \hline
\textbf{F} & 0.9858 & 0.9888 & \underline{0.9893} & 0.9893 & 0.9908 & \underline{\textbf{0.9908}} & 0.9892 & 0.9910 & \underline{\textbf{0.9925}} \\ \hline
\end{tabular}%
}
\caption{Results obtained after the fake news classification task where the results for each iteration of our approach with various deep learning classification methods are shown [P (Precision), R (Recall), and F (F-Score)]. (-) $\xrightarrow{}$ No Evidence, S1 $\xrightarrow{}$ Summarization-1 as Evidence, S2 $\xrightarrow{}$ Summarization-2 as Evidence.}
\label{tab:results}
\end{table*}

\section{Experiment Setup}
In this section, we discuss the experiment setup in detail. We perform the task of fake news detection as a binary classification task in a supervised setting. We choose to perform our experiments with both conventional machine learning- and deep learning- based classifiers. From the machine learning-based approaches, we choose Logistic Regression (LR) and Support Vector Machines (SVM) with the GridSearch implementation for best results over multiple hyperparameters (values of c, different kernels, \textit{etc.}) We also utilize LSTMs with various contextual language models from the deep learning methods. From the deep learning-based approaches, we use a simple LSTM implementation with pre-trained GloVE\footnote{https://nlp.stanford.edu/projects/glove/} vectors, BERT$_{base}$, RoBERTa$_{base}$, and XLNET$_{base}$ -based classifiers. Our LSTM implementation uses \textit{Adam} optimizer with a learning rate of 0.001, and 256 as the batch size. For classifiers based on BERT$_{base}$, RoBERTa$_{base}$, and XLNET$_{base}$, we use the HuggingFace implementations with a batch size of 32, L2 regularization and cross-entropy loss. The regularization parameter $\lambda$ was set to 0.1. Each classification method is iterated \textbf{(1)} without evidence (-), \textbf{(2)} with augmented summarized evidences from S1, \textbf{(3)} and then with S2, thus giving us three sets of results for each method; as shown in Table~\ref{tab:results}.

As an input to the classifier, we use the claim as-is from the dataset as described above. We have a dataset $D$ $= (x_n, y_n)_{n=1}^{N}$ comprising of $N$ training samples. Here $x_n = (c_n, e_n)$, where $c_n$ represents the claim, and $e_n$ represents evidence gathered using our approach. $X\in \mathcal{X}$ is defined on input space, and $Y\in\mathcal{Y} = \{0,1\}$ are the corresponding labels. Thus, given a claim $c$ and evidence $e$, the aim of this task is to train a classifier such that the claim $c$ is predicted as fake news or not, i.e $F_{\theta}:\mathcal{X} \rightarrow \mathcal{Y} \in\{0, 1\}$.
  \begin{equation}
    F(c, e;\theta)=
    \begin{cases}
      1, & \text{if}\ c \ is  \ the \ fake \ news \\
      0, & \text{otherwise}
    \end{cases}
  \end{equation}

where $F(c, e)$ is the function our model aims to learn over each iteration or epoch.

\section{Results and Discussion}

\begin{table*}[!ht]
\centering
\begin{tabularx}{\linewidth}{|>{\raggedright\arraybackslash}X|c|c|c|} \hline
\textbf{Text} & \textbf{XLNet} & \textbf{LR} & \textbf{SVM} \\ \hline
We always appreciate questions about the quality of our data. If you see a number that doesn't look right please file an issue at and we will investigate.  [SEP] SOURCES: github.com & \xmark & \checkmark & \checkmark\\\hline

The number of daily tests has been increasing in a steep climb. Average daily tests during the past three weeks also strongly depict the progress made in enhancement of \#COVID19 tests across the country.  [SEP] SOURCES: twitter.com/MoHFW\_INDI & \xmark & \checkmark & \checkmark \\\hline
Bill Gates said thousands of people will die with the COVID-19 vaccine  [SEP] SOURCES: & \xmark & \xmark & \xmark  \\ \hline
\end{tabularx}
\caption{Qualitative error analysis of some output cases both in terms of successes and failures of our approach.}
\label{table:error}
\end{table*}

The results for our classification task are shown in Table~\ref{tab:results}. Using our approach, we are able to marginally outperform (+0.24, F-Score) the previous state-of-the-art (SoTA) approaches for the task of fake news detection as shown in the last column (XLNet$_{base}$, S2). Even the RoBERT$_{base}$ model is able to outperform the SoTA approaches by a small margin. We present the values of our top two best models in boldface in Table~\ref{tab:results}. Although the improvement margin is small, we would like to note that the previous SoTA approaches are already performing at almost a 0.99 F-score. \textit{We executed our model run multiple times to ensure that our improvement margin is indeed truly obtained.} We also observe that RoBERTA$_{base}$, and XLNet$_{base}$ outperform the SoTA approaches (Chen et. al. / Li et. al.) even with S1 summarization component. Classical machine learning-based approaches are also shown to perform very well for this task as the scores of 0.96 can be considered to be a good performance for any classification method.

However, this is not the only key takeaway from these results. \textit{We observe that by using our novel approach, a consistent improvement is seen in the task results}. The efficacy of our approach can be seen from Table~\ref{tab:results}, as either S1 or S2 consistently outperforms all the base models (-) [no evidence] in the table.  Moreover, using our approach, we are able to gather key evidence for such a dataset where, to begin with, only claims were present with manually annotated labels. Table~\ref{table:error} illustrates the success and failure cases from XLNet, Logistic Regression and Support Vector Machines. We observe that first two cases were incorrectly predicted by the XLNet but were predicted correctly by LR and SVM. Last case was predicted incorrectly by all of the models. We believe that the absence of source in the text could be a potential reason for this failure. Our approach can gather the evidence using a fully automated method with summarization component(s) in the pipeline. The importance of this component can be gathered from manual observations of examples in the augmented dataset. We observe that summarized evidences shorten the length of the evidences, which helps the Transformer architecture-based classifiers like BERT$_{base}$, RoBERTa$_{base}$, XLNet$_{base}$ perform better. These pre-trained models have a token length limitation of 512 tokens which is easily able to capture our summarized evidence. We also manually observe that the summarization component helps reduce redundancy in the generated sentences and removes duplicates. Hence, improving the quality of evidence used as additional input helps reduce the training time. The performance of our models with the fine-tuned summarization component (S2) seems to perform better than S1, and the model without any evidence, as can be seen in Table~\ref{tab:results}. 

We acknowledge that the CONSTRAINT dataset is saturated in terms of possible improvements. However, with this paper, \textit{our aim is to show the efficacy of our summarization technique which can help the evidence detection for news}. We chose this dataset at an early stage of our work, and our experiments do show that improvements can, in fact, still be shown on this dataset. Our best-performing system surpasses the state-of-the-art by 0.23\% points. 

\section{Conclusion and Future Work}
In this paper, we present an automated method to collect evidence for the fake news detection task. We use our novel approach to augment the dataset, released in the CONSTRAINT-2021 Shared Task, with evidence sets collected from the web. Our method helps process these evidence sets, clean them and use them to generate summarized evidence based on two different methodologies. We use either of the summarized evidence as an additional input to the fake news classification task and perform an evaluation of our approach. We discuss the results of the classification task and conclude that our approach helps outperform the previous SoTA approaches by a small margin, however, helping generate evidence for a crucial dataset. We show that a summarization module can help collect evidence more effectively. We augment this dataset with the summarized evidence and release it along with the code and generated models for further research. We would also like to conclude that our method is generalizable; since it uses pre-trained metrics (BERTScore) and models (T5), it can be used to gather evidence for other datasets. The overall pipeline is also not very time-consuming (2 seconds per sample) once fine-tuned models are included in it. We hope our method and the resources are helpful to the NLP community.

In future, we would like to use our method to gather evidence for other fact detection/verification datasets as well. Our initial aim is to reproduce this study with other datasets and ensure that our method performs well in a real-world scenario. We would also like to apply this method and gather further evidence for existing fake news datasets, and perform our experiments to evaluate this approach over multiple exisiting datasets, including existing multilingual datasets.

\bibliography{anthology,acl2020}
\bibliographystyle{acl_natbib}



\end{document}